# KILOVOLT PYROELECTRIC VOLTAGE GENERATION AND ELECTROSTATIC ACTUATION WITH FLUIDIC HEATING

*Di Ni[1], Ved Gund[1], Landon Ivy[1], and Amit Lal[1*]*
[1]SonicMEMS Laboratory, School of Electrical and Computer Engineering, Cornell University, Ithaca, NY, USA

## ABSTRACT

Integrated micro power generators are crucial components for micro robotic platforms to demonstrate untethered operation and to achieve autonomy. Current micro robotic electrostatic actuators typically require hundreds to thousands of voltages to output sufficient work. Pyroelectricity is one such source of high voltages that can be scaled to small form factors. This paper demonstrates a distributed pyroelectric high voltage generation mechanism to power kV actuators using alternating exposure of crystals to hot and cold water ($30^0$C to $90^0$C water temperature). Using this fluidic temperature control, a pyroelectrically generated voltage of 2470 V was delivered to a 2 pF storage capacitor yielding a 6.10 µJ stored energy. A maximum energy of 17.46 µJ was delivered to a 47 pF capacitor at 861 V. The recirculating water can be used to heat a distributed array of converters to generate electricity in distant robotic actuator sections. The development of this distributed system would enable untethered micro-robot to be operated with a flexible body and free of battery recharging, which advances its applications in the real world.

## INTRODUCTION

Micro-robotic actuators, such as electrostatic, piezoelectric, and dielectric elastomer actuators (DEAs), require high voltages to operate. Recent developments in DEAs and electrostatic actuators have resulted in microrobots of 1 cm$^3$ scale operating at kVs [1], [2]. Because of these demands in high voltage amplitude, the majority of existing micro robot platforms are tethered to external power supplies [3], [4], which limits their applications in the real world for autonomous operation. Moreover, electrostatic actuators greatly benefit from operating with larger voltages due to the improved electromechanical coupling that enables them to have better efficiencies and higher force densities [1]. Yet, building compact power sources that can generate sufficient actuation voltage for micro actuators [3], [5], [6] while fulfilling the small size, weight, and power (SWaP) constraint remains to be a great challenge [3], [5], [6].

To deliver sufficient voltages to actuators mentioned above, integrated power electronic circuits are commonly used to boost voltages from a few volts to thousands of volts. Table 1 compares the existing techniques used for low-to-high voltage conversion. For insect-scale robots (0.1-1 cm$^3$), the µ-HV supply should be compact and light in order to add the least amount of extra payload, while generating 0.2-10 kV for decent operations of the actuators. Power electrical converters, such as boost and flyback converters, use solenoid inductors as energy storage units. The RoboBee robot used a custom-wound inductor with a very light weight of 5 mg, and achieved a maximum voltage of 190V and a moderate efficiency of 55% [7]. Such small inductors are also extremely hard to make and have a yield of less than 50%. Unlike these electromagnetic coupling converters where inductors are necessary, piezoelectric transformers convert AC primary voltage to a secondary voltage using electromechanical coupling with strain amplification [8], [9]. Though they have benefited from small sizes and high energy density, piezoelectric transformers generally have low voltage gain, fixed operation frequency, and suffer from large leakage at high voltages [7], [10]. Recently, switched capacitor converters have achieved HV while maintaining relatively small sizes. These converters replace inductors with capacitors for energy storage and realize voltage gain through the control of switching transistors [11]. However, these converters need specialized high-voltage CMOS circuitries for precise switching, and generally have a limited gain of ~10x due to the complexity in exact timing.

*Table 1: Comparison of existing µ-HV supplies*

| Converters | Size | Mass | Output Voltage | Efficiency |
|---|---|---|---|---|
| Flyback Converter [7], [12] | 178 cm$^3$ | 100 g | 5000 V | ~ 75% |
| | 0.1 cm$^3$ | 20 mg | 260 V | ~ 70% |
| Switched capacitor [11] | 0.12 mm$^3$ | 300 mg | 1500 V | ~ 80% |
| Piezoelectric Transformer [9], [12], [13] | < 1 mm$^2$ PCB surface | 2 g | 500 V | 50% (theoretical) |
| Pyroelectric generator (this work) | 0.2 cm$^3$ (without fluidic channel) | 570 mg | 2500 V | < 0.1% |

Pyroelectricity provides another promising approach to generating high voltages in small and lightweight form factors. Pyroelectric materials have spontaneous polarization, whose amplitude is modulated reversibly with oscillating temperatures. As a result, a pyroelectrically-generated voltage is built across the crystal surfaces. One of the most widely used pyroelectric materials, lithium niobate (LiNbO$_3$, or LN), has been demonstrated for kilovolts voltage generation within a small volume of 0.2 cm$^3$ with a simple architecture consisting of only three parts: a heater, a pyroelectric crystal, and a regulating switch [14]–[16]. Resistive heating in these systems utilizes electrical energy and can ideally be replaced by using ambient thermal energy.

In this work, we demonstrate distributed pyroelectric voltage generation and actuation with fluid-mediated heating and cooling, which serves as an alternative to the resistive heating approach reported previously [14], [15]. This system provides capabilities for scavenging the broadly existing thermal energy in the environment and convert it to electricity, and enabling untethered microrobot to operate without the need for battery recharging. An array of pyroelectric crystals is physically connected in series by being glued to the fluidic channel. The fluid delivers heat to each of the crystals, while slowly losing the heat content in the fluid. The voltage generated on each crystal can power an actuator locally, minimizing the parasitic capacitance and wiring complexity usually encountered from high-voltage compatible wiring.

This paper is arranged as follows: an introduction to the distributed pyroelectric high voltage system is first provided, followed by the equivalent thermal circuit of a unit. The experimental results on voltage stored on capacitors and actuation of an electrostatic actuator is then addressed.

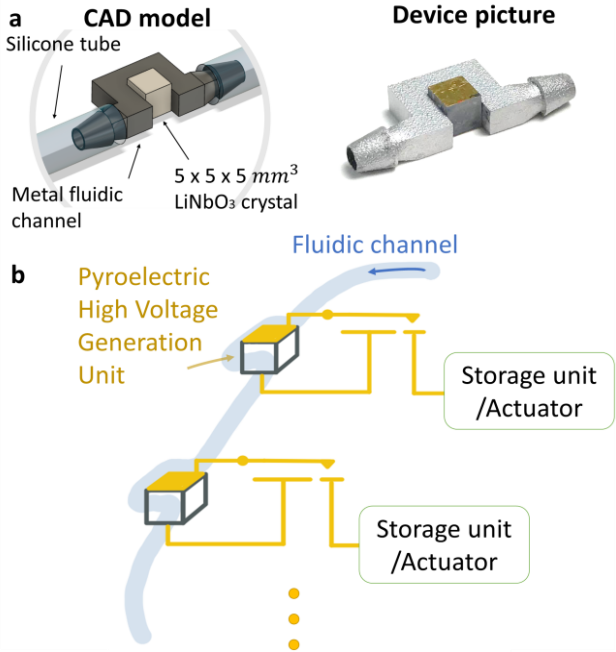

*Figure 1: Schematic of the pyroelectric high voltage generation system. **a.** Each pyroelectric crystal (5x5x5 mm³ LiNbO₃, coated with gold on the top and bottom surfaces) is surrounded with a 3D printed metal holder. **b.** The fluidic channel is used to heat and cool an array of crystals with running liquid. Charges generated on the crystals enable the local powering of actuators, or voltage delivery to a single storage capacitor.*

## DISTRIBUTED PYROELECTRIC HIGH VOLTAGE GENERATION

The pyroelectric effect is caused by the change of spontaneous polarization under the temperature variation. The amount of heat delivered to the crystal determines the quantity of charge and voltage generated on the crystal. With a temperature change of $\Delta T$, the amount of pyroelectric charge can be estimated from:

$$Q_{pyroelectric} = pA\Delta T$$

where $p$ is the pyroelectric coefficient and $A$ is the surface area of the crystal.

Due to the inevitable charge leakage within the crystal and charge-neutralization with the free-floating electrons and ions in the air, the generation rate of pyroelectric charges must exceed the leakage of charges for the voltage to steadily accrue. Thus, the rate of temperature change is also an important parameter to be considered in modeling the final voltage reachable on a pyroelectric crystal. The pyroelectric crystal operates as a current source, whose amplitude is determined by:

$$i_{pyro} = pA\frac{dT_{pyro}}{dt}$$

A KCL analysis for the circuit shown in Fig. 3a, can be used to determine the pyroelectric voltage $V_{pyro}$:

$$i_{pyro} = (C_{pyro} + C_{store} + C_{Parasitic})\frac{dV_{pyro}}{dt} + \frac{V_{pyro}}{R_{leak}}$$

and $R_{leak}$ represents equivalent resistance for charge leakage paths.

Figure 1 shows an overview of the system, consisting of the Pyroelectric High Voltage Generation (PHVG) units, and a fluidic channel used to connect 3D-printed alloy crystal holders. Each of the lithium niobate crystals serves as s pyroelectric voltage generator. Due to its low permittivity, low dielectric loss, moderate pyroelectric coefficient, and easy commercial availability of high-quality wafers, lithium niobate is an ideal material for this system. The preparation of LiNbO₃ crystals consisted of two steps: (i) a z-cut, 5 mm thick, 1-inch diameter wafer was first sputtered with Au on both surfaces; (ii) the wafer was diced into 5 x 5 mm² pieces using DISCO dicing saw. The crystal is surrounded on 3 sides by a 3D-printed metal channel made of aluminum alloy (Alsi10Mg), which was designed to have good thermal conductivity to transfer the fluid heat to the crystal. A thin layer of thermal adhesive (MG Chemicals 8329TCS) was used to fix the crystal into the 3D printed corner.

Figure 1b shows a schematic of the distributed system. Multiple PHVG units are connected in series with a fluidic channel. A peristaltic liquid pump was used to pump hot/cold liquid from the storage tank to circulate in the tube, enabling the thermal energy converted to electricity at multiple crystal sites. We used water as a preliminary test liquid, whose temperature was varied between 30⁰C and 90⁰C, with a hot plate used to heat a beaker of water which served as the recirculation reservoir. While being heated/cooled, pyroelectrically-generated voltages are produced across the surfaces of a LiNbO₃ crystal. The built-up charges are shared between the pyroelectric crystal and the capacitance of an electro-mechanical electrostatic switch connected in series with one of the crystal electrodes. The switches act as a gap-varying capacitor and experiences pull-in instability at the threshold voltage determined by the gap to the drain and actuation area. Since the heat-driven events happen with a frequency of less than 10 Hz, the current loss on the mechanical switch is very small. Once the pyroelectric voltage reaches the pull-in instability point, the mechanical switch closes and delivers the pyroelectric voltage to the load, which can be either a storage capacitor or a capacitive actuator. In the current setup, the loaded actuators are limited to capacitive actuators due to the limited charge LiNbO₃ can generate owing to a smaller surface area. For current-driven actuators, other pyroelectric crystals, such as PMN-PT and PZT, may serve as better alternatives.

The system enables high voltages to be generated at the distributed PHVG sites, limited only by the dielectric breakdown of air. Such an architecture will aid powering a robot with many actuators, such as the soft robot introduced in [17], where DEAs are employed at each bending joint of a snake robot. The key advantage of this design is the localized actuation and storage of kilovolts of voltages, which avoids the charge dissipation and arcing in charge delivery.

## EQUIVALENT THERMAL CIRCUIT MODEL

Figure 2a shows the equivalent thermal model of a PHVG unit, which accounts for the thermal resistance and capacitance of the components. The liquid was modeled as a heat source, whose temperature was assumed to remain constant during the heat transfer process. This is a reasonable assumption because, as compared to the metal alloy and the crystal, the running water has a much higher heat capacity and consequently a higher equivalent thermal capacitance and mass. The thermodynamics of a running liquid were not included in this model for simplicity. LiNbO₃ crystals were bonded with thermal adhesive to the metal channels for optimal heat transfer from the circulating fluid. The heat from the liquid needs to propagate through the metal channel and the thermal adhesive layer to reach the crystal, therefore these three modules (metal channel, thermal adhesive, crystal) are connected in series. Each module was modeled as a combination of a thermal resistor with a temperature drop, and a heat storage capacitor. The time constant of the thermal conduction can be calculated with: $\tau = RC$ where $R$ is the thermal resistance from the fluid to the crystal, and $C$ is the thermal capacitance of the crystal and the crystal holder. Both the metal alloy and the thermal adhesive have high thermal conductivities and short propagation paths (metal channel wall thickness 500 $\mu m$, thermal adhesive thickness <100 $\mu m$), heat can be transferred from the liquid to the pyroelectric crystal in less than 100 picoseconds.

Figure 2b shows the experimental measurements of the temperature change rate $dT/dt$ of a crystal heated by water at different temperatures, along with the simulation results from the equivalent circuit model. Since the series-connected thermal resistors are linear components, the model predicts a linear relationship between the heating source and the temperature change rate at the pyroelectric crystal. For each experimental point, the maximum temperature change rates at given fluidic temperatures were plotted as measured with an IR camera and a bonded resistance temperature device (RTD). The measured data shows an increase in the changing rate with increasing water temperature. A deviation from the linear model is potentially due to variation in the thermal resistance of the adhesive due to varying adhesive thickness, as well as fluctuations due to heat transfer to the RTD device and surface emissivity for IR measurements.

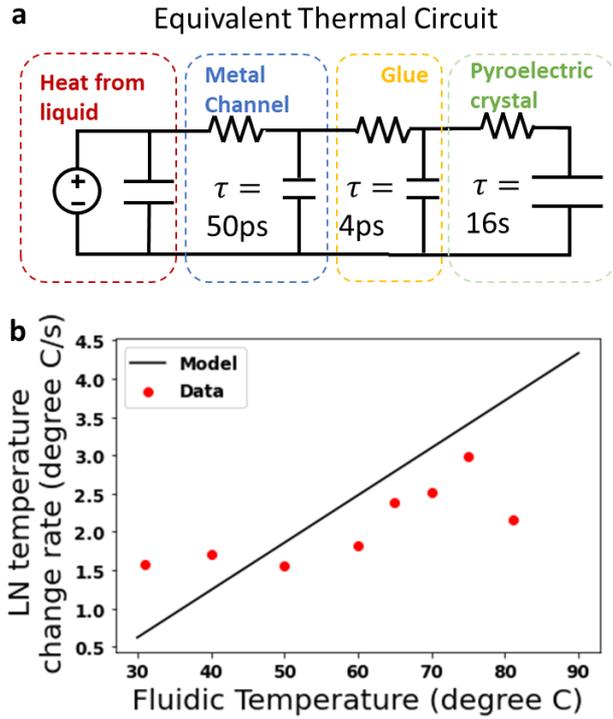

*Figure 2. Characterization of the thermal properties. **a.** An equivalent thermal circuit model consists of the metal channel, the thermal glue, and the pyroelectric crystal. The heated liquid is considered as an energy source with pulse output. **b.** The rate of temperature change happened on a LN crystal temperature is linearly proportional to the temperature of the running liquid.*

**EXPERIMENTAL RESULTS**

The PHVG system was then tested to deliver high voltages to storage capacitors. Figure 3a shows the electrical circuit used for these experiments. A PHVG unit was connected in parallel with a storage capacitor and a cantilever beam. The voltage measurement setup follows [16], where displacements of the cantilever beam connected to a LiNbO$_3$ crystal were measured with a laser and calibrated to convert the observed displacement to estimate generated voltages. This indirect measurement technique is useful because it has no charge consumption during measurement.

Three capacitors, with values of 2pF, 10pF, and 47 pF, were used as loads separately to store the voltage generated from pyroelectric crystals driven by the fluidic heat transfer. Figure 3b shows the voltage delivered to and the energy stored on three capacitors. The experiments were conducted at room temperature (25$^0$C). The larger a storage capacitor is, the more charges it needs to build the voltage up. A maximum voltage of 2.47 kV was achieved on a 2 pF storage capacitor at a fluidic temperature of 90$^0$C. Figure 3c shows the total energy $E = \frac{1}{2}CV^2$ stored on a capacitor. A maximum energy of 17.46 µJ was delivered to a 47 pF capacitor at 861 V.

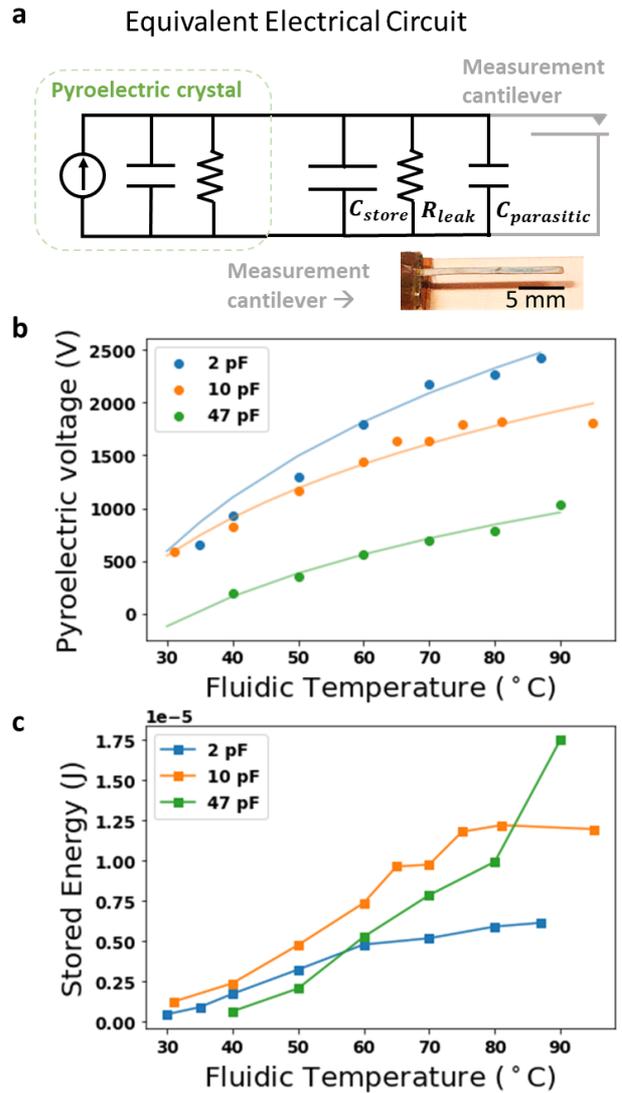

*Figure 3. Electrical characterization of the pyroelectric system. **a.** The electrical circuit of the system where a storage capacitor is connected to Lithium Niobate. A cantilever beam is used to measure voltage delivered to the capacitor. **b.** The maximum voltage delivered to three capacitors of 2, 10, and 47 pF. The solid lines represent exponential fitting of the data **c.** The maximum energy delivered to three capacitors.*

Figure 4 shows an electrostatic actuator driven by the pyroelectric high voltage generator. The actuator consists of two pieces filled with arrays of pillars, such that it can have a high surface area-to-volume ratio that results in a high energy density. A maximum displacement of 2.5 µm was achieved when actuated by the PHVG system, which corresponded to a voltage of 1033 V. This demonstration shows that PHVG system has successful fluidic control of robotic actuation.

**CONCLUSIONS**

In this work, we demonstrated a distributed high voltage generation system utilizing the pyroelectric effect and mechanical switches. A fluidic channel was used to deliver thermal energy to several high voltage generators, which presents a pathway toward a microrobotic platform with multiple actuators and a single ambient heating/cooling source. This architecture not only allows kilovolts of voltage to be generated at multiple sites but is inherently designed to make use of extra thermal energy scavenged from the environment for sustainable operation.

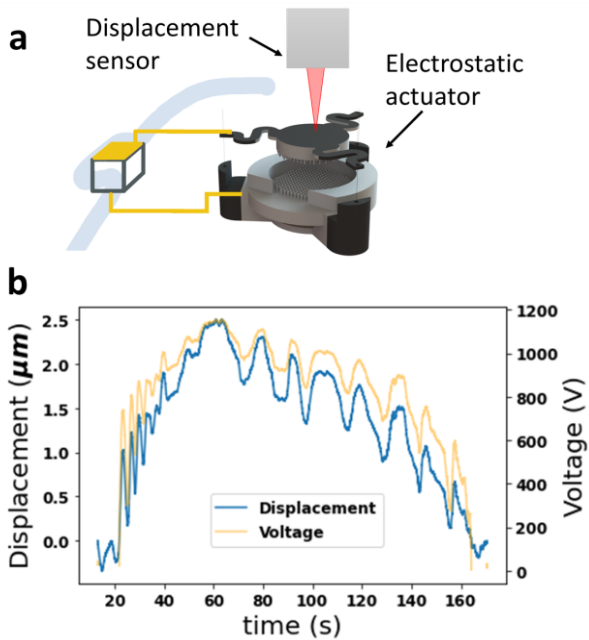

*Figure 4. Actuation of a microrobot actuator. (a) A kilovolts electrostatic actuator was driven with the pyroelectric high voltage system. The displacement of the actuator was recorded with a laser sensor. (b) The displacement performance of the electrostatic actuator.*

Further efforts are needed to simplify the setup and shrink the size, weight, and power (SWaP) of the system. For example, [18] provides an inspiration for utilizing the liquid-vapor phase transitions to induce temperature change and eliminating the use of an external pump. To drive actuators with large capacitance or high voltage requirements, multiple crystals can deliver the voltage to a single storage capacitor.

## ACKNOWLEDGEMENTS

This work is supported by the NSF EFRI program, and made use of the Cornell NanoScale Facility (NSF Grant NNCI-2025233) for device fabrication.

## CONTACT

*Di Ni; dn273@cornell.edu